\documentclass[10pt,twocolumn,letterpaper]{article}

\usepackage{cvpr}

\usepackage{multirow}
\usepackage{balance}
\usepackage{epigraph}
\usepackage{bbding}

\usepackage{tikz}

\usepackage{xcolor}
\usepackage{colortbl}
\definecolor{turquoise}{cmyk}{0.65,0,0.1,0.3}
\definecolor{purple}{rgb}{0.65,0,0.65}
\definecolor{dark_green}{rgb}{0, 0.5, 0}
\definecolor{orange}{rgb}{0.8, 0.6, 0.2}
\definecolor{red}{rgb}{0.8, 0.2, 0.2}
\definecolor{blueish}{rgb}{0.0, 0.7, 1}
\definecolor{light_gray}{rgb}{0.7, 0.7, .7}
\definecolor{pink}{rgb}{1, 0, 1}
\definecolor{dark_red}{rgb}{0.5, 0, 0}

\newcommand{\approachName}{HaMeR\xspace}
\newcommand{\dataset}{HInt\xspace}

\definecolor{hands23_color}{RGB}{151, 169, 211}
\definecolor{epick_color}{RGB}{254, 152, 103}
\definecolor{ego4d_color}{RGB}{111, 202, 174}

\usepackage{pifont}

\newcommand{\cmark}{\textcolor{green!80!black}{\ding{51}}}
\newcommand{\xmark}{\textcolor{red}{\ding{55}}}

\definecolor{cvprblue}{rgb}{0.21,0.49,0.74}

\def\paperID{****}
\def\confName{CVPR}
\def\confYear{2024}
    
\title{Reconstructing Hands in 3D with Transformers}

\author{
\kern-7mm Georgios Pavlakos$^{1}$, Dandan Shan$^{2}$, Ilija Radosavovic$^{1}$, Angjoo Kanazawa$^{1}$, David Fouhey$^{3}$, Jitendra Malik$^{1}$\\
\kern-7mm $^{1}$UC Berkeley, $^{2}$University of Michigan, $^{3}$New York University
}

\begin{document}

\twocolumn[{
\renewcommand\twocolumn[1][]{#1}
\maketitle
\begin{center}
    \vspace{-0.26in}
    \centerline{
    \includegraphics[width=\linewidth]{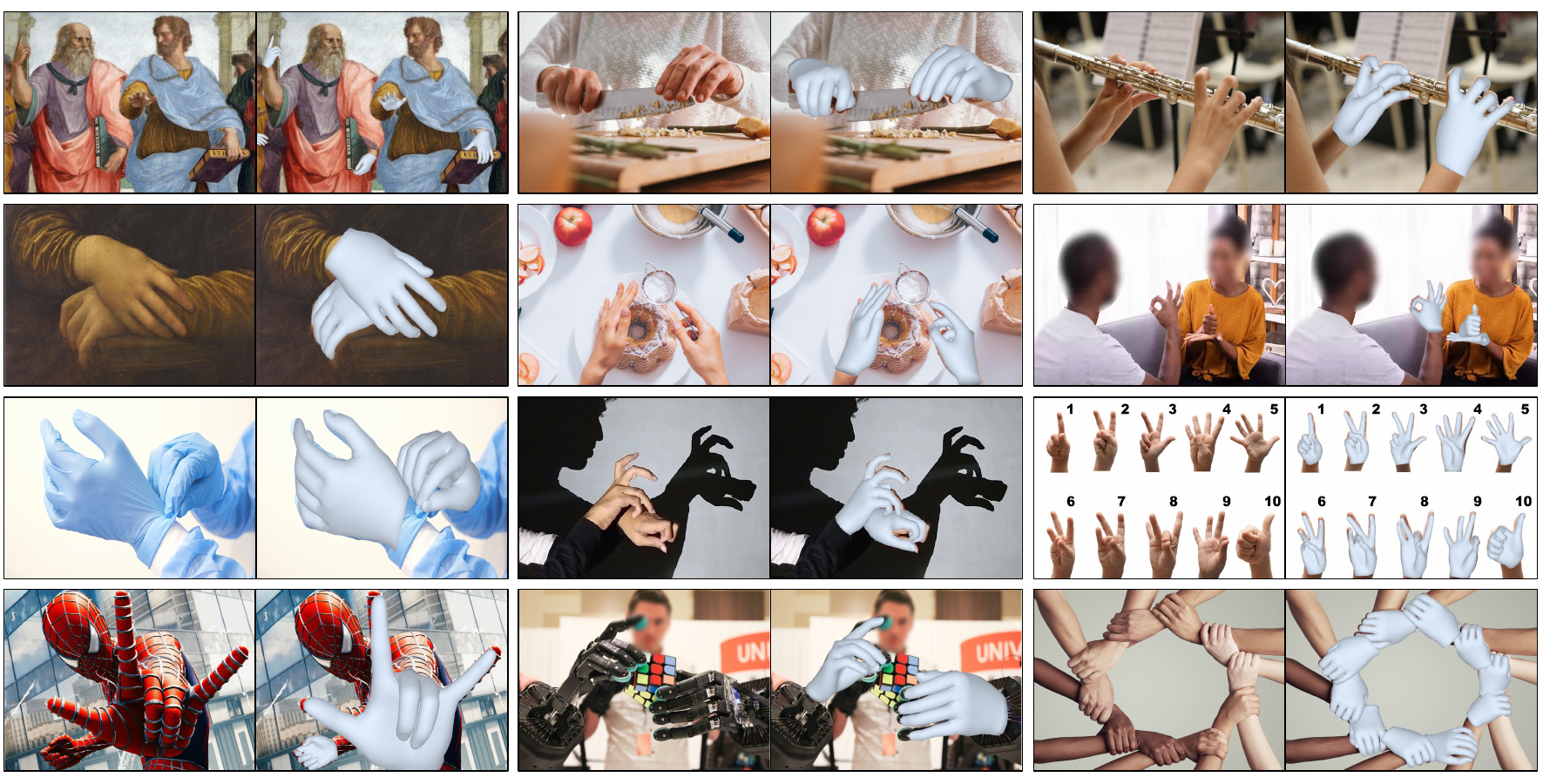}
     }
    \vspace{-0.08in}
   \captionof{figure}{{\bf Monocular 3D hand mesh reconstruction.}
   We propose \approachName, a fully %
   transformer-based approach for {\bf Ha}nd {\bf Me}sh {\bf R}ecovery.
   \approachName achieves consistent improvements upon the state-of-the-art for 3D hand reconstruction.
   We can faithfully reconstruct hands in a wide variety of scenarios,
   including captures from different viewpoints (third person or egocentric),
   under occlusion,
   hands that interact with objects or other hands,
   hands with different skin tones,
   with gloves,
   from art paintings
   or mechanical hands.
   We encourage the reader to watch our reconstructions in the Supplemental Video to appreciate the temporal stability.
   }
   \vspace{-0.02in}
\label{fig:teaser}
\end{center}
}]

\begin{abstract}
We present an approach that can reconstruct hands in 3D from monocular input.
Our approach for Hand Mesh Recovery, \approachName, follows a fully transformer-based architecture and can analyze hands with significantly increased accuracy and robustness compared to previous work.
The key to \approachName's success lies in scaling up both the data used for training and the capacity of the  deep network for hand reconstruction.
For training data, we combine multiple datasets that contain 2D or 3D hand annotations.
For the deep model, we use a large scale Vision Transformer architecture.
Our final model consistently outperforms the previous baselines on popular 3D hand pose benchmarks.
To further evaluate the effect of our design in non-controlled settings, we annotate existing in-the-wild datasets with 2D hand keypoint annotations.
On this newly collected dataset of annotations, \dataset, we demonstrate significant improvements over existing baselines.
We make our code, data and models available on the project website:
\url{https://geopavlakos.github.io/hamer/}.

\end{abstract}

\vspace{-1.0em}
\setlength{\epigraphwidth}{0.32\textwidth}
\epigraph{``It is because of his being armed with hands that man is the most intelligent animal.''}{\textit{Anaxagoras}}
\vspace{-1.8em}    
\section{Introduction}
\label{sec:introduction}

Consider the images of hands interacting with the world in Figure~\ref{fig:teaser}.
These interactions are happening in 3D, so to interpret them,
we also need a system that can automatically perceive hands in 3D from visual input.

Recent developments 
in computer vision and NLP 
point to the direction where advances are achieved by simple, high capacity models, powered by huge amounts of data.
This emerging insight has been demonstrated in the context of NLP by Large Language Models, like GPT-3~\cite{brown2020language} and \mbox{GPT-4}~\cite{openai2023gpt4}.
In the context of computer vision, we observe this with models like CLIP~\cite{radford2021learning}, Stable Diffusion~\cite{rombach2022high} and SAM~\cite{kirillov2023segment}.
In the area of 3D human mesh recovery, a similar trend has been observed, where the simple, large scale HMR2.0 architecture~\cite{goel2023humans} achieves state-of-the-art results.

In this paper, we take this philosophy and apply it to the problem of 3D hand pose estimation.
We propose \approachName, a robust and accurate approach for {\bf Ha}nd {\bf Me}sh {\bf R}ecovery from images and video frames.
\approachName captures faithful 3D reconstructions of hands in a variety of poses, viewpoints and visual conditions, as shown in Figure~\ref{fig:teaser}.
This translates to improvements over existing baselines in the standard 3D hand pose benchmarks.
More importantly, \approachName shines when evaluated on challenging in-the-wild images, where we outperform the state-of-the-art by significant margins.
Even though \approachName is a single-frame approach, it recovers temporally smooth and consistent reconstructions when applied on video frames (please see the Supplemental Video for video results).

The key to \approachName's success lies in scaling up the techniques for hand mesh recovery.
More specifically, we scale both the training data and the deep network architecture used for 3D hand reconstruction.
For training data, we use multiple available sources of data with hand annotations, including  both studio/controlled datasets with 3D ground truth~\cite{chao2021dexycb,hampali2020honnotate,moon2020interhand2,xiang2019monocular,zimmermann2017learning,zimmermann2019freihand}, and in-the-wild datasets annotated with 2D keypoint locations~\cite{fang2022alphapose,jin2020whole,simon2017hand}.
For our network, we use a large-scale transformer architecture~\cite{dosovitskiy2020image,xu2022vitpose} which can successfully consume data of this scale.
The combination of these two ingredients leads to  significant improvements compared to previous work. %

Benchmarking progress of these models is challenging and is often constrained on datasets captured in controlled conditions.
To encourage evaluation on in-the-wild images, we introduce a new dataset of annotations, \dataset, %
by annotating hands from diverse image sources, including videos from YouTube~\cite{shan2020understanding,cheng2023towards} and egocentric captures~\cite{damen2018scaling, grauman2022ego4d}.
The annotations consist of 2D keypoints for the hand joints, as well as labels of the visibility (occluded or not) for each joint.
We provide 2D hand keypoints annotations for 40.4K hands, where 86.7\% of them are hands in natural contact. %
Even though with \dataset we can only benchmark the 2D aspect of our 3D reconstruction, 
this evaluation is complementary to the existing benchmarks due to its diversity of data, and together provide a more holistic picture on the performance of different systems.

We contribute \approachName, an approach for 3D hand mesh reconstruction from images and video frames.
We demonstrate the key effect of scaling up to large scale training data and high capacity deep architectures for the problem of hand mesh recovery.
We achieve state-of-the-art results where we obtain 2-3$\times$ improvement in PCK@0.05 on in-the-wild datasets compared to previous works.
We also contribute \dataset, a dataset of annotations that complements training and evaluation of 3D hand reconstruction approaches.
We make our model, code and data available to support future work. 
\section{Related work}
\label{sec:related_work}

\noindent
{\bf 3D hand pose and shape estimation.}
In this section we focus specifically on the works that estimate 3D hand pose and shape from a single RGB image.
The earlier efforts~\cite{boukhayma20193d, baek2019pushing, zhang2019end} take inspiration from related work on human mesh recovery~\cite{kanazawa2018end} - they use the MANO parametric hand model~\cite{romero2017embodied} and regress the hand pose and shape parameters given an RGB image as input.
FrankMocap~\cite{rong2021frankmocap} is a good representative of this line of works which adopts a simple design, similar to HMR~\cite{kanazawa2018end}.
Followup work~\cite{choi2020pose2mesh,ge20193d,kulon2020weakly,moon2020i2l} follows a non-parametric approach and directly regresses the vertices of the MANO mesh.
This strategy often leads to results that align better with the image evidence, but it is more prone to failure in cases of occlusions and truncations.
The improvements in 3D hand pose estimation have also lead to progress in related problems, including joint hand pose and object reconstruction~\cite{hasson2019learning, hasson2020leveraging, tse2022collaborative, yang2022artiboost} and reconstruction of two interacting hands~\cite{kim2021end,li2022interacting,meng20223d,moon2023bringing,ren2023decoupled,wang2023memahand,yu2023acr,zhang2021interacting,zuo2023reconstructing}.
More recently, there have been works that address other aspects of the problem.
MobRecon~\cite{chen2022mobrecon} focuses on high inference speed, that could potentially be supported on a mobile device.
HandOccNet~\cite{park2022handoccnet} designs an architecture that could offer increased robustness to occlusions.
AMVUR~\cite{jiang2023probabilistic} proposes a probabilistic approach for hand pose and shape estimation.
BlurHand~\cite{oh2023recovering} focuses on the problem of motion blur that often exists in footage that captures hand motion.
Our work is orthogonal to these approaches.
We adopt a simple design and we investigate the effect of scaling up the training data and the capacity of our architecture.
Given that our main design is simple, the different choices of previous work could be combined with our \approachName architecture which could potentially lead to further improvements.

\begin{figure*}[!h]
    \centering
    \small
    \includegraphics[width=\linewidth]{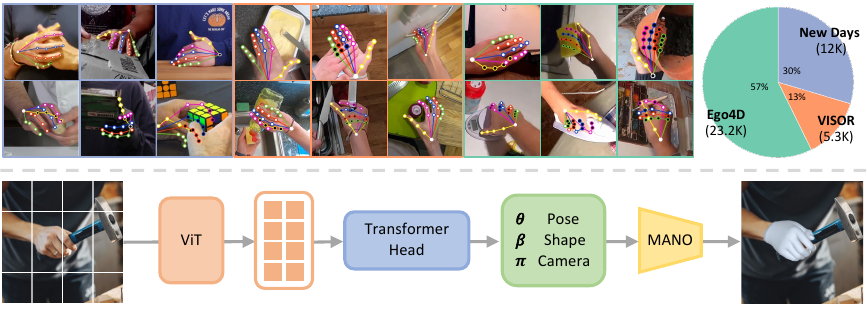}
    \caption{{
    \bf Dataset and Architecture.} {\bf (Top)} Hand crops with keypoint annotations from our \dataset dataset of annotations for different image sources, \textcolor{hands23_color}{Hands23}~\cite{cheng2023towards}, \textcolor{epick_color}{Epic-Kitchens}~\cite{VISOR2022, damen2018scaling}, and \textcolor{ego4d_color}{Ego4D}~\cite{grauman2022ego4d}. We provide location annotations for 21 hand keypoints as well as the ``occlusion'' label for each joint. 
    Occluded keypoints are marked using solid dot filled with black while non-occluded ones are filled with white.
    The pie chart shows the distribution and statistics of our dataset. {\bf (Bottom)} The architecture for \approachName follows a fully transformer-based design.
    We use a large scale ViT backbone~\cite{dosovitskiy2020image} followed by a transformer decoder to regress the parameters of the hand.
    }
    \vspace{-1.5em}
    \label{fig:dataset_arch}
\end{figure*}

\noindent
{\bf Hand datasets.}
Many of the datasets used to train and evaluate 3D hand pose estimation systems are captured in indoor/studio settings and provide 3D ground truth.
FreiHAND~\cite{zimmermann2019freihand} is captured in a multi-camera setting and focuses on different hand poses as well as hands interacting with objects.
HO-3D~\cite{hampali2020honnotate} and DexYCB~\cite{chao2021dexycb} are also captured in a controlled setting with multiple cameras but focuses more specifically on cases where hands interact with objects.
InterHand2.6M~\cite{moon2020interhand2} is captured in a studio with a focus on two interacting hands.
Hand pose datasets~\cite{simon2017hand, xiang2019monocular} captured in the Panoptic studio~\cite{joo2015panoptic} also offer 3D hand annotations.
AssemblyHands~\cite{ohkawa2023assemblyhands} annotated 3D hand poses for synchronized images from Assembly101~\cite{sener2022assembly101} which participants assemble and disassemble take-apart toys in a multi-camera setting. 
In this work, we use these datasets for training and evaluation.
However, we also argue that to get a more holistic picture about the accuracy and the robustness of 3D hand pose estimation systems, it is important to evaluate performance on in-the-wild images as well.

While we cannot annotate 3D ground truth poses for in-the-wild images, there is work that annotates 2D keypoint positions.
Among the larger scale efforts, COCO-WholeBody~\cite{jin2020whole} provides hand annotations for the people in the COCO dataset~\cite{lin2014microsoft} and Halpe~\cite{fang2022alphapose} annotates hands in the HICO-DET dataset~\cite{chao2015hico,chao2018learning}. 
Both of them source images from image datasets that contain very few egocentric images or transitionary moments.  
In our dataset, \dataset, we sourced images from both egocentric and third-person video datasets.
Since our annotated hands come from video, they depict more natural interactions with the world. 
\section{Technical approach}
\label{sec:technical_approach}

In this section, we describe \approachName, our approach for hand mesh recovery from monocular input.
We follow a simple, fully ``transformerized'' design that focuses on scaling up the training data and the deep model architecture.

\subsection{MANO parametric hand model}
We adopt the MANO parametric model of the human hand~\cite{romero2017embodied}.
MANO takes as input the pose parameters $\theta \in \mathbb{R}^{48}$ and shape parameters $\beta \in \mathbb{R}^{10}$ and defines a function $\mathcal{M}(\theta, \beta)$ that returns the mesh of the hand $M \in \mathbb{R}^{V \times 3}$, with $V=778$ vertices.
MANO additionally returns the joints $X \in \mathbb{R}^{K \times 3}$ of the hand, for a total of $K=21$ joints.

\subsection{Hand mesh recovery}
Given an RGB image of a hand, $I$, our goal is to reconstruct the 3D hand surface.
We approach this problem by estimating the MANO pose and shape parameters for the hand in the image.
Similar to previous work in the parametric human~\cite{kanazawa2018end,goel2023humans} and hand~\cite{rong2021frankmocap,zhang2019end} reconstruction, we use a network to learn the mapping $f$ from image pixels to MANO 
parameters.
Our regressor also estimates camera parameters $\pi$.
The camera $\pi$ corresponds to a translation $t \in \mathbb{R}^3$ that allows us to project the 3D mesh and the 3D joints to the image.
Given fixed camera intrinsics $K$, the projection of the 3D joints $X$ is:  $x = \pi(X) = \Pi_K(X+t)$.
Eventually, we learn the mapping $f(I) = \Theta$, where the regressed parameters are $\Theta = \{\theta, \beta, \pi\}$.

\subsection{Architecture}
\approachName adopts a simple architecture
with a fully transformer-based design (Figure~\ref{fig:dataset_arch}, bottom), similar to~\cite{goel2023humans}.
We use a Vision Transformer (ViT)~\cite{dosovitskiy2020image} as the backbone,
followed by a transformer head that regresses the hand and camera parameters.
We first convert the input RGB image to patches, which are fed as input tokens to ViT which follows the ``huge'' design, \ie, ViT-H.
The ViT backbone processes the image patches and returns a series of output tokens.
The transformer head is a transformer decoder that processes a single token while cross-attending to the ViT output tokens.
The output of the head returns the parameters $\Theta$ for the input image.
  
\subsection{Losses}
For our training losses, we follow best practices for parametric human and hand reconstruction~\cite{goel2023humans, kanazawa2018end, kolotouros2019learning, rong2021frankmocap} and supervise our model with a combination of 2D and 3D losses.
For the images that provide 3D ground truth, we can directly apply a loss on the model parameters, $\theta$ and $\beta$.
Simultaneously, we can encourage consistency in the actual 3D space, and supervise on the level of the 3D joints $X^*$:
\begin{equation}
\mathcal{L}_\texttt{3D} = ||\theta - \theta^*||_2^2 + ||\beta - \beta^*||_2^2 + ||X - X^*||_1.
\end{equation}

To enable training with 2D annotations, we also apply a reprojection loss between the projection $x$ of the 3D joints~$X$ and the ground truth 2D keypoint annotations $x^*$:
\begin{equation}
\mathcal{L}_\texttt{2D} = ||x - x^*||_1.
\end{equation}
We apply this loss, even when 3D ground truth is available, since it promotes consistency on the output image space.

Finally, if only 2D keypoints are available, it is possible to recover an unnatural pose that still reprojects well to the image.
To encourage the reconstruction of natural hands, we train discriminators $D_k$ for a) the hand shape $\beta$, b) the hand pose $\theta$, and c) each hand joint angle separately~\cite{kanazawa2018end}.
Then, we can apply an adversarial loss:
\begin{equation}
\mathcal{L}_\texttt{adv} = \sum_k(D_k(\Theta) - 1)^2.
\end{equation}

\subsection{Training data}
To train our model, we consolidate multiple datasets that provide 2D or 3D hand annotations.
Specifically, we use
FreiHAND~\cite{zimmermann2019freihand}, HO3D~\cite{hampali2020honnotate},
MTC~\cite{xiang2019monocular}, 
RHD~\cite{zimmermann2017learning}, InterHand2.6M~\cite{moon2020interhand2}, H2O3D~\cite{hampali2020honnotate},
DEX YCB~\cite{chao2021dexycb},
COCO WholeBody~\cite{jin2020whole},
Halpe~\cite{fang2022alphapose} and
MPII NZSL~\cite{simon2017hand}.
This results to 2.7M training examples,
which is $4\times$ larger than the training set of the popular FrankMocap system~\cite{rong2021frankmocap}.
The majority of this data is collected in controlled environments (\eg, studio or multi-camera setup), while only 5\% of the training examples (COCO WholeBody, Halpe and
MPII NZSL) include images from in-the-wild datasets.
\section{\dataset: \underline{H}and \underline{Int}eractions in the wild}
\label{sec:hands_dataset}
In this section, we describe the dataset we contribute, with the goal to complement existing datasets used for training and evaluation.
Since we focus on {\bf H}and {\bf Int}eractions in the wild, we call our dataset \dataset.
\dataset annotates 2D hand keypoint locations and occlusion labels for each keypoint.
We built off of Hands23~\cite{cheng2023towards} (using an early copy otained from the authors), Epic-Kitchens~\cite{damen2018scaling}, and Ego4D~\cite{grauman2022ego4d}.

By sourcing from video datasets, we harvest more transitional moments and natural poses, compared with sourcing from image data.
For \dataset, we source frames from three video datasets. In Hands23, we choose from the New Days subset~\cite{cheng2023towards} containing YouTube video frames of humans engaging in daily activities. In Epic-Kitchens, we choose frames from VISOR~\cite{VISOR2022} containing frames extracted from cooking actions. In Ego4D~\cite{grauman2022ego4d}, we choose frames from the critical frames (pre45, pre30, pre15, pre-frame, contact-frame, point-of-no-return frame, and post-frame) in the FHO (Forecasting Hands and Objects) task.

For our validation and test set, we randomly sample frames to keep data distribution the same as source datasets. For the training set, our goal is to include more challenging samples to compensate for other existing 2D keypoints datasets. Thus, for New Days and VISOR, we chose half of the samples using random sampling and forcing the other half to contain hand-object or hand-hand interaction. For Ego4D, we still randomly sample frames since the critical frames already typically focus on interactions.

Annotating hand keypoints from scratch can be time-consuming. Similar to~\cite{jin2020whole}, we initialize the annotation procedure with an existing keypoint detection model~\cite{mmpose2020} to get rough keypoint locations.
Given the annotation instructions (details in the Supplemental Material), workers are asked to correct the keypoint locations (see annotation samples in Figure~\ref{fig:dataset_arch}, top).
Additionally, each keypoint is annotated with an ``existence'' and an ``occlusion'' label. Existence indicates whether the keypoint exists within the image frame or not. Occlusion indicates whether the keypoint is occluded or not. To the best of our knowledge, \dataset is the first dataset to provide ``occlusion'' annotations for 2D hand keypoints. We believe this can lead to a more fine-grained analysis of the pose estimation systems.

In total, we annotate 40.4K hands with keypoints, 12.0K for New Days, 5.3K for VISOR, and 23.2K for Ego4D.
In our Ego4D subset, we annotate 9.3K hands from sequences, which could help future evaluation of temporal tasks.

Finally, we perform an annotation consistency check, by having 90 valid images annotated twice.
Across this subset, 90.5\% of the occlusion labels and 100\% of ``existence'' labels are consistent. In terms of keypoint locations, 94.6\% visible keypoints have offset distance within $0.25\times$ of the palm length
(see details about the data annotation process and analysis in the Supplemental Material).
\section{Experiments}
\label{sec:experiments}
In this section, we present the quantitative and qualitative evaluation of our system.
First, we evaluate the 3D pose accuracy (subsection~\ref{sec:eval3d}) and the 2D pose accuracy (subsection~\ref{sec:eval2d}) of \approachName.
Then, we ablate some characteristics of our system (subsection~\ref{sec:analysis}) and present qualitative results and comparisons (subsection~\ref{sec:qualitative}).

\subsection{3D pose accuracy}
\label{sec:eval3d}

To evaluate the 3D accuracy of \approachName, we use two standard benchmarks for 3D hand pose estimation, FreiHAND~\cite{zimmermann2019freihand}
and HO3Dv2~\cite{hampali2020honnotate}.
Both datasets are collected in controlled multi-camera environments and provide 3D ground truth annotations in the form of 3D hand meshes (using the MANO model).
To be comparable with previous work, we follow the typical protocols~\cite{lin2021mesh,park2022handoccnet}, and we report metrics that evaluate 3D joint and 3D mesh accuracy.
These metrics include PA-MPJPE and $\textrm{AUC}_\textrm{J}$ (3D joints evaluation), PA-MPVPE, $\textrm{AUC}_\textrm{V}$, F@5mm and F@15mm (3D mesh evaluation).

We present the complete results for FreiHAND in Table~\ref{tab:freihand} and for HO3Dv2 in Table~\ref{tab:ho3d}.
We compare with many baselines that estimate the 3D hand mesh from a single image in parametric or non-parametric form (\ie, regressing hand model parameters or hand model vertices respectively).
We observe that our \approachName approach achieves state-of-the-art results and consistently outperforms the previous work across the majority of the metrics.

\subsection{2D pose accuracy}
\label{sec:eval2d}

Although the 3D hand pose datasets provide accurate 3D ground truth for evaluation, they are typically collected in controlled settings, which limits the variety of subjects, viewpoints, objects of interactions, environments, etc.
To better analyze the properties of the different hand pose estimation systems, we also propose to evaluate on our \dataset benchmark that is closer to real in-the-wild conditions, compared to the previous 3D benchmarks.
The annotations of \dataset are in the form of 2D keypoints.
Metrics based on 2D only evaluate reprojection accuracy of 3D methods, but due to the nature of the images (\ie, in the wild), we can get complementary evidence about the performance of our method.
For evaluation, we report results with the commonly used PCK metric~\cite{yang2012articulated}, computed at different thresholds.
Given the form of \dataset, we provide a more detailed analysis, reporting separate results for images coming from New Days~\cite{cheng2023towards}, VISOR~\cite{damen2018scaling} and Ego4D~\cite{grauman2022ego4d}.
Moreover, we provide more fine-grained results, considering all the joints, considering only the joints that have been annotated as visible (non-occluded), or considering only the joints that have been annotated as occluded.

The complete results are presented in Table~\ref{tab:pck}.
Here, we compare with a number of recent 3D hand mesh estimation approaches that provide publicly available code.
Similarly with the results on FreiHAND and HO3D, we observe that our method outperforms the previous baselines.
However, on these datasets we observe much larger improvements. %
This highlights the clear improvement in the robustness of our approach which performs consistently across a variety of benchmarks.
Performance on FreiHAND and HO3D tends to be more saturated and it is not surprising that the margin of improvement for our approach on these datasets is smaller.
In contrast, performance on in-the-wild datasets is more representative of the robustness of the approaches in different visual conditions, different viewpoints and different interactions, \eg, contacts with surrounding objects.

\addtolength{\tabcolsep}{-5pt}
\begin{table}[!t]
  \resizebox{\columnwidth}{!}{
  \centering
  \small
\begin{tabular}{@{}lcccc@{}}
\toprule
{\footnotesize Method} & {\footnotesize PA-MPJPE $\downarrow$} & {\footnotesize PA-MPVPE $\downarrow$} & {\footnotesize F@5 $\uparrow$} & {\footnotesize F@15 $\uparrow$} \\ \midrule
I2L-MeshNet~\cite{moon2020i2l} & 7.4 & 7.6 & 0.681 & 0.973 \\
Pose2Mesh~\cite{choi2020pose2mesh} & 7.7 & 7.8 & 0.674 & 0.969 \\
I2UV-HandNet~\cite{chen2021i2uv} & 6.7 & 6.9 & 0.707 & 0.977 \\
METRO~\cite{lin2021end} & 6.5 & 6.3 & 0.731 & 0.984 \\
Tang~\etal~\cite{tang2021towards} & 6.7 & 6.7 & 0.724 & 0.981 \\
Mesh Graphormer~\cite{lin2021mesh} & 5.9 & 6.0 & 0.764 & 0.986 \\ 
MobRecon~\cite{chen2022mobrecon} & {\bf 5.7} & 5.8 & 0.784 & 0.986 \\
AMVUR~\cite{jiang2023probabilistic} & 6.2 & 6.1 & 0.767 & 0.987 \\ \midrule
Ours & 6.0 & {\bf 5.7} & {\bf 0.785} & {\bf 0.990} \\ \bottomrule
\end{tabular}
}
\vspace{-0.5em}
    \caption{\textbf{Comparison with the state-of-the-art on the FreiHAND dataset~\cite{zimmermann2019freihand}.} We use the standard protocol and
    report metrics for evaluation of 3D joint and 3D mesh accuracy. PA-MPVPE and PA-MPJPE numbers are in mm.}
    \vspace{-0.5em}
  \label{tab:freihand}%
\end{table}%
\addtolength{\tabcolsep}{5pt}

\addtolength{\tabcolsep}{-5pt}
\begin{table}[!t]
\resizebox{\columnwidth}{!}{
  \centering
  \scriptsize
\begin{tabular}{@{}lcccccc@{}}
\toprule
Method & $\textrm{AUC}_\textrm{J}$ $\uparrow$ & PA-MPJPE $\downarrow$ & $\textrm{AUC}_\textrm{V}$ $\uparrow$ & PA-MPVPE $\downarrow$ & F@5 $\uparrow$ & F@15 $\uparrow$ \\ \midrule
Liu~\etal~\cite{liu2021semi} & 0.803 & 9.9 &0.810&9.5&0.528&0.956\\
HandOccNet~\cite{park2022handoccnet} & 0.819 & 9.1 &0.819&8.8&0.564&0.963\\
I2UV-HandNet~\cite{chen2021i2uv} & 0.804 & 9.9 &0.799&10.1&0.500&0.943\\
Hampali~\etal~\cite{hampali2020honnotate} & 0.788 & 10.7 &0.790&10.6&0.506&0.942\\
Hasson~\etal~\cite{hasson2019learning} & 0.780 & 11.0 &0.777&11.2&0.464&0.939\\
ArtiBoost~\cite{yang2022artiboost} & 0.773 & 11.4 &0.782&10.9&0.488&0.944\\
Pose2Mesh~\cite{choi2020pose2mesh} & 0.754 & 12.5 &0.749&12.7&0.441&0.909\\
I2L-MeshNet~\cite{moon2020i2l} & 0.775 & 11.2 &0.722&13.9&0.409&0.932\\
METRO~\cite{lin2021end} & 0.792 & 10.4 & 0.779 & 11.1 & 0.484 & 0.946\\
MobRecon\cite{chen2022mobrecon}& -&9.2&-& 9.4& 0.538& 0.957\\
Keypoint Trans~\cite{hampali2022keypoint} & 0.786 & 10.8 &-&-&-&-\\
AMVUR~\cite{jiang2023probabilistic} & 0.835 & 8.3 & 0.836 & 8.2 & 0.608 & 0.965 \\ \midrule
Ours & {\bf 0.846} & {\bf 7.7} & {\bf 0.841} & {\bf 7.9} & {\bf 0.635} & {\bf 0.980} \\ \bottomrule
\end{tabular}
}
\vspace{-0.7em}
    \caption{\textbf{Comparison with the state-of-the-art on the HO3D dataset~\cite{hampali2020honnotate}.} We use the HO3Dv2 protocol and report metrics that evaluate accuracy of the estimated
    3D joints and 3D mesh. PA-MPVPE and PA-MPJPE numbers are in mm.}
    \vspace{-1.5em}
  \label{tab:ho3d}%
\end{table}%
\addtolength{\tabcolsep}{5pt}

\begin{table*}[!h]
  \centering
  \small
\begin{tabular}{@{}c|lccccccccccc@{}}
\toprule
\multirow{2}{*}{} &\multicolumn{1}{c}{\multirow{2}{*}{Method}} & \multicolumn{3}{c}{New Days} &  & \multicolumn{3}{c}{VISOR} &  & \multicolumn{3}{c}{Ego4D} \\
& \multicolumn{1}{c}{}                        & @0.05        & @0.1     & @0.15     &  & @0.05          & @0.1     & @0.15               &  & @0.05          & @0.1     & @0.15          \\ \midrule
\parbox[t]{2mm}{\multirow{6}{*}{\rotatebox[origin=c]{90}{\color{dark_green} \bf All Joints}}} & FrankMocap~\cite{rong2021frankmocap} & 16.1 & 41.4 & 60.2 & & 16.8 & 45.6 & 66.2 & & 13.1 & 36.9 & 55.8 \\
& METRO~\cite{lin2021end} & 14.7 & 38.8 & 57.3 & & 16.8 & 45.4 & 65.7 & & 13.2 & 35.7 & 54.3 \\
& MeshGraphormer~\cite{lin2021mesh} & 16.8 & 42.0 & 59.7 & & 19.1 & 48.5 & 67.4 & & 14.6 & 38.2 & 56.0 \\
& HandOccNet (param)~\cite{park2022handoccnet} & 9.1 & 28.4 & 47.8 & & 8.1 & 27.7 & 49.3 & & 7.7 & 26.5 & 47.7 \\
& HandOccNet (no param)~\cite{park2022handoccnet} & 13.7 & 39.1 & 59.3 & & 12.4 & 38.7 & 61.8 & & 10.9 & 35.1 & 58.9 \\
& Ours & {\bf 48.0} & {\bf 78.0} & {\bf 88.8} & & {\bf 43.0} & {\bf 76.9} & {\bf 89.3} & & {\bf 38.9} & {\bf 71.3} & {\bf 84.4} \\ 
\midrule
\parbox[t]{2mm}{\multirow{6}{*}{\rotatebox[origin=c]{90}{\color{dark_green} \bf Visible Joints}}} & FrankMocap~\cite{rong2021frankmocap} & 20.1 & 49.2 & 67.6 & & 20.4 & 52.3 & 71.6 & & 16.3 & 43.2 & 62.0 \\
& METRO~\cite{lin2021end} & 19.2 & 47.6 & 66.0 & & 19.7 & 51.9 & 72.0 & & 15.8 & 41.7 & 60.3 \\
& Mesh Graphormer~\cite{lin2021mesh} & 22.3 & 51.6 & 68.8 & & 23.6 & 56.4 & 74.7 & & 18.4 & 45.6 & 63.2 \\
& HandOccNet (param)~\cite{park2022handoccnet} & 10.2 & 31.4 & 51.2 & & 8.5 & 27.9 & 49.8 & & 7.3 & 26.1 & 48.0 \\
& HandOccNet (no param)~\cite{park2022handoccnet} & 15.7 & 43.4 & 64.0 & & 13.1 & 39.9 & 63.2 & & 11.2 & 36.2 & 60.3 \\
& Ours & {\bf 60.8} & {\bf 87.9} & {\bf 94.4} & & {\bf 56.6} & {\bf 88.0} & {\bf 94.7} & & {\bf 52.0} & {\bf 83.2} & {\bf 91.3} \\ 
\midrule
\parbox[t]{2mm}{\multirow{6}{*}{\rotatebox[origin=c]{90}{\color{dark_green} \bf Ocluded Joints}}} & FrankMocap~\cite{rong2021frankmocap} & 9.2 & 28.0 & 46.9 & & 11.0 & 33.0 & 55.0 & & 8.4 & 26.9 & 45.1 \\
& METRO~\cite{lin2021end} & 7.0 & 23.6 & 42.4 & & 10.2 & 32.4 & 53.9 & & 8.1 & 26.2 & 44.7 \\
& MeshGraphormer~\cite{lin2021mesh} & 7.9 & 25.7 & 44.3 & & 10.9 & 33.3 & 54.1 & & 8.3 & 26.9 & 44.6 \\
& HandOccNet (param)~\cite{park2022handoccnet} & 7.2 & 23.5 & 42.4 & & 7.4 & 26.1 & 46.7 & & 8.0 & 26.1 & 45.7 \\
& HandOccNet (no param)~\cite{park2022handoccnet} & 9.8 & 31.2 & 50.8 & & 9.9 & 33.7 & 55.4 & & 9.6 & 31.1 & 52.7 \\
& Ours & {\bf 27.2} & {\bf 60.8} & {\bf 78.9} & & {\bf 25.9} & {\bf 60.8} & {\bf 80.7} & & {\bf 23.0} & {\bf 56.9} & {\bf 76.3} \\ \bottomrule
\end{tabular}
\vspace{-0.5em}
    \caption{\textbf{Evaluation on our \dataset benchmark.} We report results using PCK scores at three different thresholds. All methods are 3D and we evaluate the scores through the 2D projection of 3D joints. We report separate results for the three subsets of \dataset, \ie, New Days of Hands~\cite{cheng2023towards}, Epic- Kitchens VISOR~\cite{VISOR2022} and Ego4D~\cite{grauman2022ego4d}. We also report separate results considering all joints (first set of rows), considering only the joints annotated as visible (second set of rows), or considering only the joints annotated as occluded (third set of rows).}
    \vspace{-1.0em}
  \label{tab:pck}%
\end{table*}%

\subsection{Ablation analysis}
\label{sec:analysis}

Having demonstrated the effectiveness of \approachName,
we further ablate different options for our system.

\addtolength{\tabcolsep}{-5pt}
\begin{table}[!h]
\resizebox{\columnwidth}{!}{
  \centering
  \footnotesize
\begin{tabular}{@{}c|lccccccccccc@{}}
\toprule
\multirow{2}{*}{} &\multicolumn{1}{c}{\multirow{2}{*}{Method}} & \multicolumn{3}{c}{New Days} &  & \multicolumn{3}{c}{VISOR} &  & \multicolumn{3}{c}{Ego4D} \\
& \multicolumn{1}{c}{}                        & @0.05        & @0.1     & @0.15     &  & @0.05          & @0.1     & @0.15               &  & @0.05          & @0.1     & @0.15          \\ \midrule
\parbox[t]{2.5mm}{\multirow{2}{*}{\rotatebox[origin=c]{90}{\color{dark_green} \bf \footnotesize All}}} & Ours & 48.0 & 78.0 & 88.8 & & 43.0 & 76.9 & 89.3 & & 38.9 & 71.3 & 84.4 \\
& Ours$^*$ & {\bf 51.6} & {\bf 81.9} & {\bf 91.9} & & {\bf 56.5} & {\bf 88.1} & {\bf 95.6} & & {\bf 46.9} & {\bf 79.3} & {\bf 90.4} \\ 
\midrule
\parbox[t]{2.5mm}{\multirow{2}{*}{\rotatebox[origin=c]{90}{\color{dark_green} \bf \footnotesize Vis.}}} & Ours & 60.8 & 87.9 & 94.4 & & 56.6 & 88.0 & 94.7 & & 52.0 & 83.2 & 91.3 \\
& Ours$^*$ & {\bf 62.9} & {\bf 89.4} & {\bf 95.8} & & {\bf 66.5} & {\bf 92.7} & {\bf 97.4} & & {\bf 59.1} & {\bf 87.0} & {\bf 94.0} \\ 
\midrule
\parbox[t]{2.5mm}{\multirow{2}{*}{\rotatebox[origin=c]{90}{\color{dark_green} \bf \footnotesize Occl.}}} & Ours & 27.2 & 60.8 & 78.9 & & 25.9 & 60.8 & 80.7 & & 23.0 & 56.9 & 76.3 \\
& Ours$^*$ & {\bf 33.2} & {\bf 68.4} & {\bf 84.8} & & {\bf 42.6} & {\bf 79.0} & {\bf 91.3} & & {\bf 33.1} & {\bf 69.8} & {\bf 84.9} \\ \bottomrule
\end{tabular}
}
\vspace{-0.8em}
    \caption{\textbf{Effect of training with \dataset.}
    We compare our general model (Ours) with the model trained on \dataset as well (Ours$^*$).
    We report PCK scores on the test set of \dataset.
    Using the training set of \dataset can be helpful particularly to improve performance on egocentric data (VISOR and Ego4D).}
    \vspace{-1.5em}
  \label{tab:finetuning}%
\end{table}%
\addtolength{\tabcolsep}{5pt}

\noindent
{\bf Effect of large scale data and deep model.}
One of the key aspects of \approachName is that a simple design can achieve strong performance if we scale up, \ie, train with large scale data and use a large scale model for the hand reconstruction.
We evaluate these choices using different models on \dataset and we present the complete results in Table~\ref{tab:design}.
More specifically, we start from a basic design (2nd row of Table~\ref{tab:design}), that follows the choices of~\cite{rong2021frankmocap} (1st row of Table~\ref{tab:design}), using a ResNet50 architecture~\cite{he2016deep} and a relatively small training set (only a quarter of the examples we use to train \approachName).
This basic design is indeed very close to~\cite{rong2021frankmocap} in terms of quantitative results.
Then, by keeping the architecture the same, we increase the volume of training examples, using our complete training set.
This model (3rd row of Table~\ref{tab:design}) achieves already consistent improvements over the previous baseline.
Similarly, if we use the small training set of the basic design, but adopt a large scale architecture, here \mbox{ViT-H}~\cite{dosovitskiy2020image} (4th row of Table~\ref{tab:design}), we also see improvements over the basic design.
Finally, we can combine the two independent updates, \ie, increase the volume of training examples while using a high capacity architecture, which effectively is the design of \approachName.
This version (5th row of Table~\ref{tab:design}) outperforms by a large margin the other versions, demonstrating the effect of both large data and large deep model in our design.

\begin{table*}[!h]
\resizebox{\textwidth}{!}{
  \centering
  \small
\begin{tabular}{@{}c|lccccccccccccc@{}}
\toprule
\multirow{2}{*}{} &\multicolumn{1}{c}{\multirow{2}{*}{Method}} & Large & Large & \multicolumn{3}{c}{New Days} &  & \multicolumn{3}{c}{VISOR} &  & \multicolumn{3}{c}{Ego4D} \\
& \multicolumn{1}{c}{}                & Data &  Model       & @0.05        & @0.1     & @0.15     &  & @0.05          & @0.1     & @0.15               &  & @0.05          & @0.1     & @0.15          \\ \midrule
\parbox[t]{2.5mm}{\multirow{5}{*}{\rotatebox[origin=c]{90}{\color{dark_green} \bf \footnotesize All}}} & FrankMocap~\cite{rong2021frankmocap} & \xmark & \xmark & 16.1 & 41.4 & 60.2 & & 16.8 & 45.6 & 66.2 & & 13.1 & 36.9 & 55.8 \\
& Base design & \xmark & \xmark & 16.9 & 43.6 & 62.6 & & 17.5 & 47.5 & 67.3 & & 13.9 & 37.8 & 56.0 \\
& \quad + large data &  \cmark & \xmark & 31.3 & 65.7 & 81.8 & & 29.9 & 65.0 & 81.7 & & 24.7  & 56.1 & 73.9 \\
& \quad + large model & \xmark & \cmark & 25.9 & 58.9 & 76.9 & & 24.1 & 62.5 & 81.2 & & 19.4 & 51.6 & 71.1 \\
& \approachName & \cmark & \cmark & {\bf 48.0} & {\bf 78.0} & {\bf 88.8} & & {\bf 43.0} & {\bf 76.9} & {\bf 89.3} & & {\bf 38.9} & {\bf 71.3} & {\bf 84.4} \\ 
\midrule
\parbox[t]{2.5mm}{\multirow{5}{*}{\rotatebox[origin=c]{90}{\color{dark_green} \bf \footnotesize Visible}}} & FrankMocap~\cite{rong2021frankmocap} & \xmark & \xmark & 20.1 & 49.2 & 67.6 & & 20.4 & 52.3 & 71.6 & & 16.3 & 43.2 & 62.0 \\
& Base design & \xmark & \xmark & 21.2 & 51.5 & 70.4 & & 21.4 & 54.5 & 73.5 & & 17.4 & 45.0 & 63.7 \\
& \quad + large data & \cmark & \xmark & 38.5 & 75.0 & 88.0 & & 36.6 & 73.2 & 86.9 & & 30.4 & 64.8 & 80.9 \\
& \quad + large model & \xmark & \cmark & 33.1 & 69.3 & 85.0 & & 29.2 & 72.8 & 88.9 & & 24.3 & 62.3 & 81.3 \\
& \approachName & \cmark & \cmark & {\bf 60.8} & {\bf 87.9} & {\bf 94.4} & & {\bf 56.6} & {\bf 88.0} & {\bf 94.7} & & {\bf 52.0} & {\bf 83.2} & {\bf 91.3} \\ 
\midrule
\parbox[t]{2.5mm}{\multirow{5}{*}{\rotatebox[origin=c]{90}{\color{dark_green} \bf \footnotesize Occluded}}} & FrankMocap~\cite{rong2021frankmocap} & \xmark & \xmark & 9.2 & 28.0 & 46.9 & & 11.0 & 33.0 & 55.0 & & 8.4 & 26.9 & 45.1 \\
& Base design & \xmark & \xmark & 9.4 & 29.8 & 48.8 & & 11.8 & 35.6 & 57.4 & & 9.2 & 27.4 & 44.8 \\
& \quad + large data & \cmark & \xmark & 19.0 & 49.4 & 70.7 & & 19.1 & 51.6 & 72.5 & & 17.0 & 44.5 & 64.3 \\
& \quad + large model & \xmark & \cmark & 14.7 & 41.6 & 63.2 & & 16.3 & 47.9 & 69.4 & & 14.5 & 40.1 & 59.9 \\
& \approachName & \cmark & \cmark & {\bf 27.2} & {\bf 60.8} & {\bf 78.9} & & {\bf 25.9} & {\bf 60.8} & {\bf 80.7} & & {\bf 23.0} & {\bf 56.9} & {\bf 76.3} \\ \bottomrule
\end{tabular}
}
    \caption{\textbf{Effect of large scale data and deep model.}
    We evaluate the effect of different design choices when testing on \dataset.
    We start from a basic design that follows FrankMocap~\cite{rong2021frankmocap}, using a ResNet50 architecture and a small training set (2nd row).
    Increasing the amount of training data by $4\times$ (3rd row) or adopting a high capacity ViT-H architecture (4th row) results in clear and consistent improvements in 2D accuracy over the base model.
    Combining the data scale and high capacity architecture, which is the proposed \approachName (5th row), obtains the best results by large margins.
    }
  \label{tab:design}%
\end{table*}%

\noindent
{\bf Training with \dataset.}
When comparing with previous work, we avoided training with the training set of \dataset.
However, here we provide a direct comparison when training with this data too.
In Table~\ref{tab:finetuning} we present the detailed results on \dataset.
We observe a clear improvement on VISOR and Ego4D, the two egocentric datasets included in \dataset.
This can be explained by the fact that there have been little to no egocentric data with hand annotations in the wild before, so using some form of annotations for training can help improve our model.
Besides this, we also observe an improvement in New Days.
The improvement is smaller, given that New Days mainly includes third-person videos, but it is still consistent across all metrics.

\begin{figure*}[!h]
    \centering
    \small
    \includegraphics[width=1.0\linewidth]{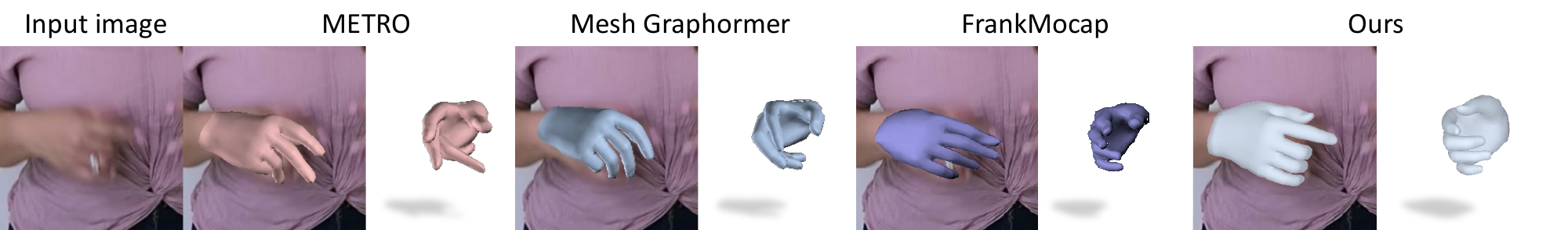}
    \includegraphics[width=1.0\linewidth]{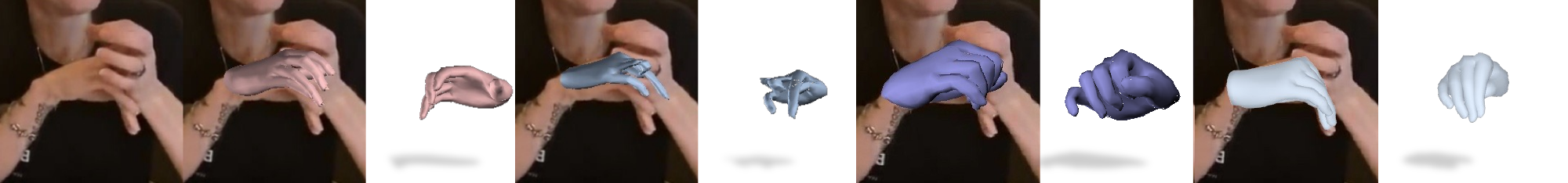}
    \includegraphics[width=1.0\linewidth]{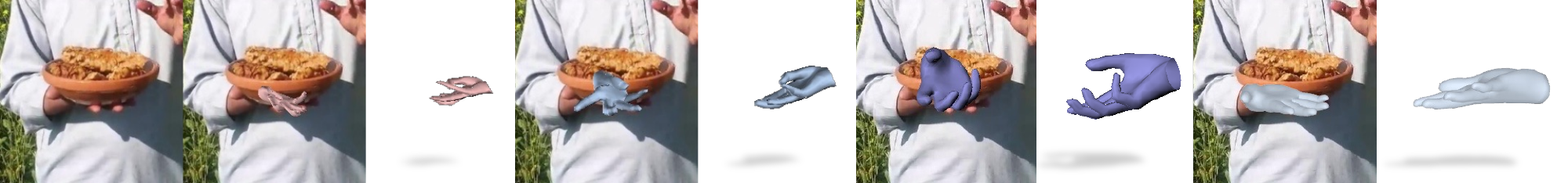}
    \includegraphics[width=1.0\linewidth]{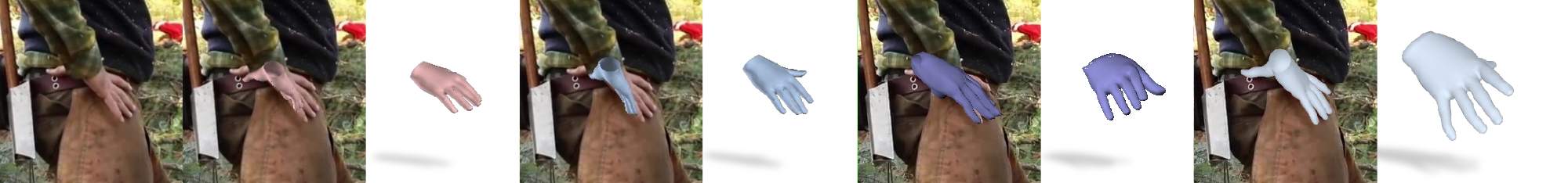}
    \caption{{\bf Qualitative comparison.} We compare our approach qualitatively with state-of-the-art methods for hand mesh reconstruction.
    The previous baselines include METRO~\cite{lin2021end}, Mesh Graphormer~\cite{lin2021mesh} and FrankMocap~\cite{rong2021frankmocap}.
    METRO and Mesh Graphormer are non-parametric methods (regressing MANO vertices directly),
    while FrankMocap and \approachName (ours) are parametric methods (regressing MANO parameters).
    The reconstructions from
    \approachName are consistently better, particularly on more challenging examples, \eg, cases with motion blur, or images with hand-hand or hand-object interaction.
    We encourage the reader to also watch the Supplemental Video for more comparisons over time. %
    }
    \label{fig:qualitatitve_comparison}
    \vspace{-1.8em}
\end{figure*}

\subsection{Qualitative results}
\label{sec:qualitative}

\begin{figure*}[!h]
    \centering
    \small
    \includegraphics[width=1.0\linewidth]{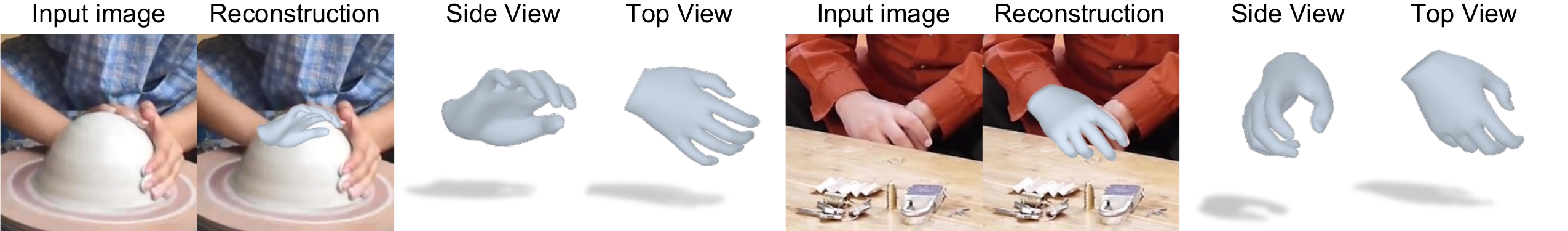}
    \includegraphics[width=1.0\linewidth]{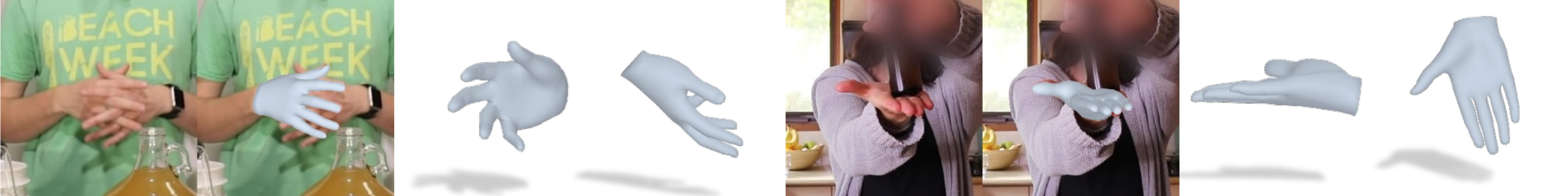}
    \includegraphics[width=1.0\linewidth]{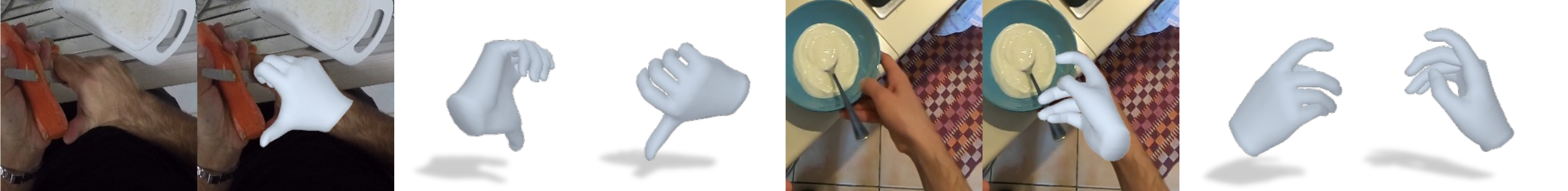}
    \includegraphics[width=1.0\linewidth]{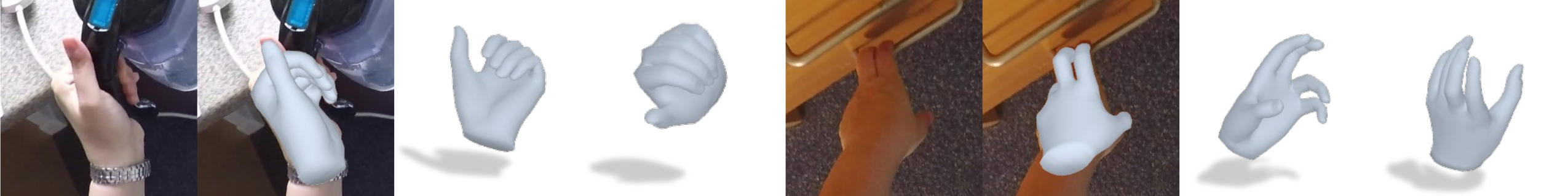}
    \includegraphics[width=1.0\linewidth]{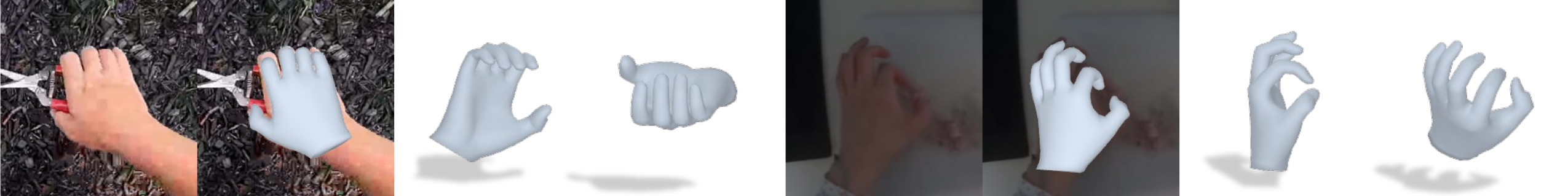}
    \includegraphics[width=1.0\linewidth]{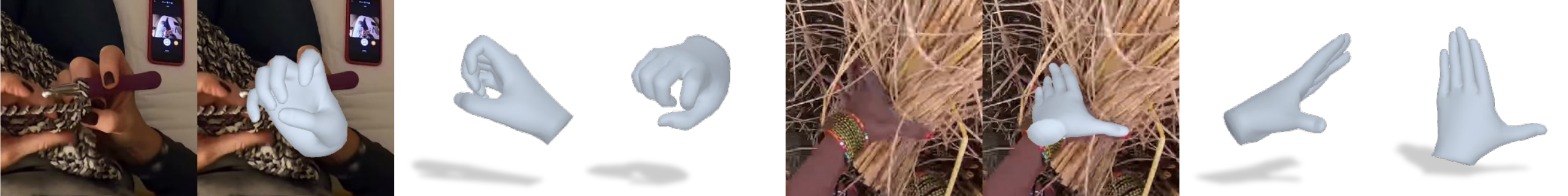}
    \includegraphics[width=1.0\linewidth]{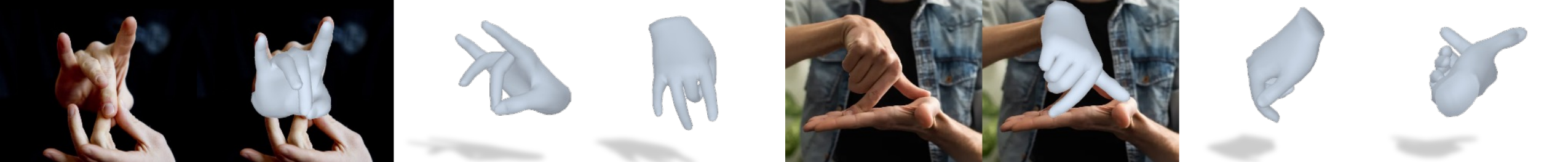}
    \includegraphics[width=1.0\linewidth]{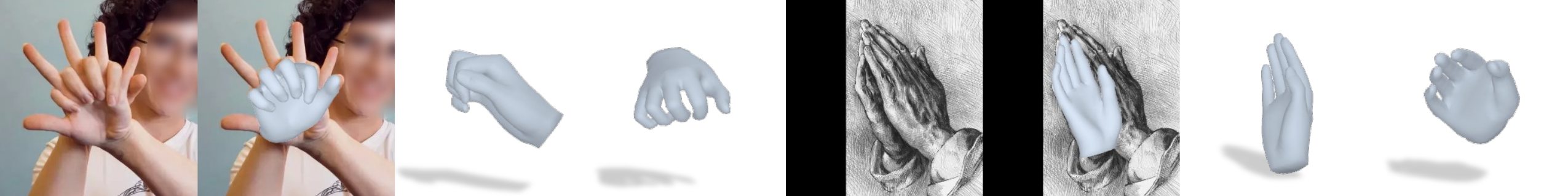}
    \caption{{\bf Qualitative results.} We present qualitative results of our approach on the test set of \dataset. We include images from New Days (row 1-2), VISOR (row 3-4), Ego4D (row 5-6), as well as various Internet images (row 7-8). \approachName is particularly robust and can gracefully handle cases with heavy occlusion and interactions with objects or other hands.
    }
    \label{fig:qualitatitve_ours}
\end{figure*}

We show qualitative results of our approach in Figure~\ref{fig:teaser}, while we do a more detailed analysis in Figure~\ref{fig:qualitatitve_ours}, where we show side and top views of our 3D hand reconstructions.
Our approach is robust to different viewpoints, different skin tones or hand appearance (\eg wearing different types of gloves) as well as different objects of interaction that can create various degrees of occlusion.
Moreover, in Figure~\ref{fig:qualitatitve_comparison}, we show more detailed comparisons with previous baselines.
We compare with METRO~\cite{lin2021end}, MeshGraphormer~\cite{lin2021mesh} and FrankMocap~\cite{rong2021frankmocap}.
Following the trend of the quantitative comparison, \approachName is consistently more robust and precise than the previous work.

\section{Conclusion}
\label{sec:conclusion}

We present \approachName, an approach for 3D hand mesh reconstruction from monocular input.
\approachName is simple, without bells and whistles and demonstrates the importance of two design choices --- scaling up the hand mesh recovery models in terms of a) the training data and b) the architecture we use for 3D hand reconstruction.
By consolidating multiple datasets with hand annotations (either 2D or 3D) and adopting a high capacity deep model (ViT-H~\cite{dosovitskiy2020image}), we are able to outperform previous work on traditional 3D hand pose benchmarks.
Additionally, we contribute 2D keypoint annotations for datasets with diverse hands, coming from egocentric~\cite{damen2018scaling,grauman2022ego4d} views or YouTube videos~\cite{cheng2023towards}.
Evaluation on this challenging new \dataset benchmark demonstrates the even bigger improvements that our approach achieves compared to previous baselines.
We hope that the robustness and the precision of our approach will ignite the interest for further use of our system in applications that 3D hand estimation is important, including, but not limited to, robotics, action recognition and sign language understanding.

\paragraph{Acknowledgements} We thank members of the BAIR community for helpful discussions and StabilityAI for supporting us through a compute grant. This work
was supported by BAIR/BDD sponsors, ONR MURI (N00014-21-1-2801), and the DARPA MCS program. DF and DS were supported by the National Science Foundation under Grant No. 2006619.

{
    \small
    \bibliographystyle{ieeenat_fullname}
    \bibliography{main}
}

\end{document}